\pdfoutput=1
\documentclass[11pt]{article}
\usepackage[letterpaper,left=1.25in,right=1.25in,top=1.15in,bottom=1.15in]{geometry}
\usepackage{times}
\usepackage{latexsym}
\usepackage[T1]{fontenc}
\usepackage[utf8]{inputenc}
\usepackage{microtype}
\usepackage{inconsolata}
\usepackage{graphicx}
\usepackage{amsmath,amssymb}
\usepackage{booktabs}
\usepackage{tabularx}
\usepackage{array}
\usepackage{multirow}
\usepackage{xcolor}
\usepackage{xspace}
\usepackage{tikz}
\usepackage{fvextra}
\usepackage[font=small,labelfont=bf]{caption}
\usepackage{natbib}
\usetikzlibrary{arrows.meta,positioning,fit,calc}

\definecolor{linkblue}{RGB}{0,60,140}
\usepackage[breaklinks=true,colorlinks=true,linkcolor=linkblue,
            citecolor=linkblue,urlcolor=linkblue]{hyperref}


\newcommand{\id}[1]{\texttt{#1}}
\newcommand{\BR}{\textsc{BlueprintRepair}\xspace}
\newcommand{\BT}{\textsc{BlueprintTrace}\xspace}
\newcolumntype{Y}{>{\raggedright\arraybackslash}X}
\newif\ifshowtodos
\showtodosfalse

\newlength{\figcol}
\title{BlueprintRepair: Typed Local Edits for\\Failed Lean Proof Blueprints}
\author{Ruslan Khrulev\\
  Lomonosov Moscow State University\\
  \texttt{ra.khrulev@gmail.com}}
\date{}

\hypersetup{%
  pdftitle={BlueprintRepair: Typed Local Edits for Failed Lean Proof Blueprints},
  pdfauthor={Ruslan Khrulev},
  pdfkeywords={formal theorem proving, Lean 4, proof repair, proof blueprints,
               large language models, tool use}}

\begin{document}
\maketitle

\begin{abstract}
LLM-based Lean proving systems increasingly organize a proof as a
\emph{blueprint}: a dependency graph of formal statements. We introduce \BR, a
repair interface that lets a model change this graph through ten
schema-checked local operations. An operation names the node it edits, so the
target theorem cannot be changed. Lean checks every applied change, and an
accepted repair must declare every blueprint lemma its proof uses. We also
construct \BT, a benchmark of $142$ controlled failures with complete accepted
and rejected repair trajectories. We compare typed edits, exact source
patches, and complete module rewrites under matched source, feedback, model,
and budget, one episode per state and interface. With DeepSeek-V4-Flash, the
three interfaces solve almost the same number of the benchmark's localized
failures. Typed repair is the cheapest per solved state (patching is
$1.30\times$ as expensive, rewriting $2.06\times$), and within $10{,}000$
completion tokens per task it reaches almost all of its final coverage, while
both free-form interfaces are well behind. A second model, Qwen3.6-Flash,
solves fewer states but keeps typed repair cheapest, puts it ahead on the
proof-authoring states, and repeats the localized pattern.
\end{abstract}

\section{Introduction}
\label{sec:intro}

LLM-based theorem provers have traditionally generated tactics or complete
proofs \citep{leandojo,baldur,dsproverv2}. Newer systems first build a
structured proof plan. In Lean \citep{lean4}, this plan can be represented as a
\emph{blueprint}: a dependency graph whose nodes are formal statements and
whose edges record which statements a proof is meant to use. LeanArchitect
maintains this metadata inside Lean developments \citep{leanarchitect};
Goedel-Architect and LeanMarathon use evolving blueprints to coordinate longer
proving and formalization processes \citep{goedelarchitect,leanmarathon}.

Once the proof plan is explicit, repair need not begin by regenerating the
whole Lean file. A failed blueprint may contain a false intermediate lemma, a
missing dependency, or an unused node. In these cases, a small change to the
existing graph may be enough. A different failure occurs when all statements
and edges are correct, but one lemma still lacks a proof. We ask:
\textbf{when are typed local edits sufficient, and when is freer Lean code
generation more useful?}

\begin{figure}[t]
\centering
\resizebox{0.98\textwidth}{!}{%
\begin{tikzpicture}[
  font=\scriptsize,
  >={Stealth[length=1.7mm]},
  node/.style={rounded corners=2pt,draw,semithick,minimum width=15mm,
    minimum height=7.4mm,inner sep=2pt,align=center},
  proved/.style={node,draw=green!45!black,fill=green!12},
  bad/.style={node,draw=red!70!black,fill=red!12},
  open/.style={node,draw=black!55,fill=black!5},
  hard/.style={node,draw=orange!80!black,fill=orange!18},
  deadn/.style={node,draw=black!45,dashed,fill=black!3,text=black!60},
  title/.style={font=\small\bfseries},
  hdr/.style={font=\scriptsize},
  callout/.style={rounded corners=2pt,draw,align=center,inner sep=3pt,
    fill=white,font=\scriptsize},
  note/.style={font=\tiny,text=red!60!black},
]
\begin{scope}[shift={(0,0)}]
\node[hdr] at (2.3,1.52) {\id{p159} (miniF2F \id{amc12b\_2003\_p9}):};
\node[hdr] at (2.3,1.14)
  {$f(x){=}ax{+}b$,\ $f(6){-}f(2){=}12\;\vdash\;f(12){-}f(2){=}30$};
\node[title] at (2.3,0.68) {(a) localized graph defect};
\node[proved] (adf) at (1.0,0)
  {{\tiny\ttfamily diff\_four}\\ $f(6){-}f(2)=4a$};
\node[proved] (adt) at (3.6,0)
  {{\tiny\ttfamily diff\_ten}\\ $f(12){-}f(2)=10a$};
\node[proved] (aav) at (1.0,-1.4)
  {{\tiny\ttfamily a\_value}\\ slope: $a=3$};
\node[bad] (afs) at (3.6,-1.4)
  {{\tiny\ttfamily final\_step}\\ $\forall a\colon 10a{=}30$\; \textbf{false}};
\node[open] (atg) at (2.3,-2.8)
  {{\tiny\ttfamily target}\\ unproved};
\draw[->] (adf) -- (aav);
\draw[->] (aav) -- (atg);
\draw[->] (afs) -- (atg);
\draw[->] (adt.south east) .. controls +(0.8,-0.95) and +(1.05,0.25) .. (atg.east);
\node[callout,draw=red!60!black,text width=46mm,anchor=north] at (2.3,-3.72)
  {injected defect: the hypothesis $a{=}3$ is removed;
   the weakened lemma is false and Lean refutes it};
\end{scope}

\begin{scope}[shift={(5.45,0)}]
\node[hdr] at (2.5,1.52) {\id{p144} (miniF2F \id{imo\_1959\_p1}):};
\node[hdr] at (2.5,1.14) {$\gcd(21n{+}4,\,14n{+}3)=1$ for $0<n$};
\node[title] at (2.5,0.68) {(b) compound state: chained defects};
\node[proved] (bgl) at (1.1,0)
  {{\tiny\ttfamily gcd\_left}\\ $g \mid 21n{+}4$};
\node[bad] (bco) at (3.7,0)
  {{\tiny\ttfamily combo}\\ $3(14n{+}3)-2(21n{+}4)=2$\\
   \textcolor{red!60!black}{\tiny false: the true combination gives $1$}};
\node[proved] (bgr) at (1.1,-1.5)
  {{\tiny\ttfamily gcd\_right}\\ $g \mid 14n{+}3$};
\node[bad] (bdv) at (3.7,-1.5)
  {{\tiny\ttfamily dvd\_one}\\ $d\mid 21n{+}4 \Rightarrow d\mid 1$\\
   \textcolor{red!60!black}{\tiny premise $d\mid 14n{+}3$ dropped}};
\node[deadn] (bsc) at (1.1,-3.0)
  {{\tiny\ttfamily small\_case}\\ $\gcd(25,17)=1$ {\tiny(debris)}};
\node[open] (btg2) at (3.55,-3.0)
  {{\tiny\ttfamily target}\\ unproved};
\draw[->] (bgl) -- (btg2);
\draw[->] (bgr) -- (btg2);
\draw[->] (bco) -- (bdv);
\draw[->] (bdv) -- (btg2);
\node[callout,draw=red!60!black,text width=48mm,anchor=north] at (2.5,-3.72)
  {a false certificate feeds a divisibility step that
   lost a premise: two defects chained on the used
   path, plus a disconnected spot-check};
\end{scope}

\draw[black!30,dotted,thick] (10.7,1.7) -- (10.7,-4.9);

\begin{scope}[shift={(10.95,0)}]
\node[hdr] at (2.15,1.52) {\id{p170} (miniF2F):};
\node[hdr] at (2.15,1.14) {the three-variable AM--HM bound};
\node[title] at (2.15,0.68) {(c) proof-authoring failure};
\node[proved] (dam) at (2.15,0)
  {{\tiny\ttfamily amhm\_cleared}\\ $9pqr\le(p{+}q{+}r)(pq{+}qr{+}rp)$};
\node[proved] (dpp) at (1.0,-1.4)
  {{\tiny\ttfamily pair\_pos}\\ pair sums $>0$};
\node[proved] (dps) at (3.35,-1.4)
  {{\tiny\ttfamily pair\_sum}\\ sum $=2(x{+}y{+}z)$};
\node[hard] (dtg) at (2.15,-2.8)
  {{\tiny\ttfamily target}\\
   $\tfrac{9}{x+y+z}\le\tfrac{2}{x+y}+\tfrac{2}{y+z}+\tfrac{2}{z+x}$};
\draw[->] (dam) -- (dtg);
\draw[->] (dpp) -- (dtg);
\draw[->] (dps) -- (dtg);
\node[callout,draw=orange!75!black,text width=40mm,anchor=north] at (2.15,-3.72)
  {every statement and edge is correct;\\ repair must supply proof content};
\end{scope}
\end{tikzpicture}}
\caption{Three benchmark examples. (a) A localized defect weakens a
lemma until it becomes false. (b) A compound state contains two linked defects
on the target path and one disconnected node. (c) In a proof-authoring state,
the graph is correct but the remaining theorem still needs proof content.
Statements are abbreviated; Table~\ref{tab:families} lists all failure
families.}
\label{fig:regimes}
\end{figure}
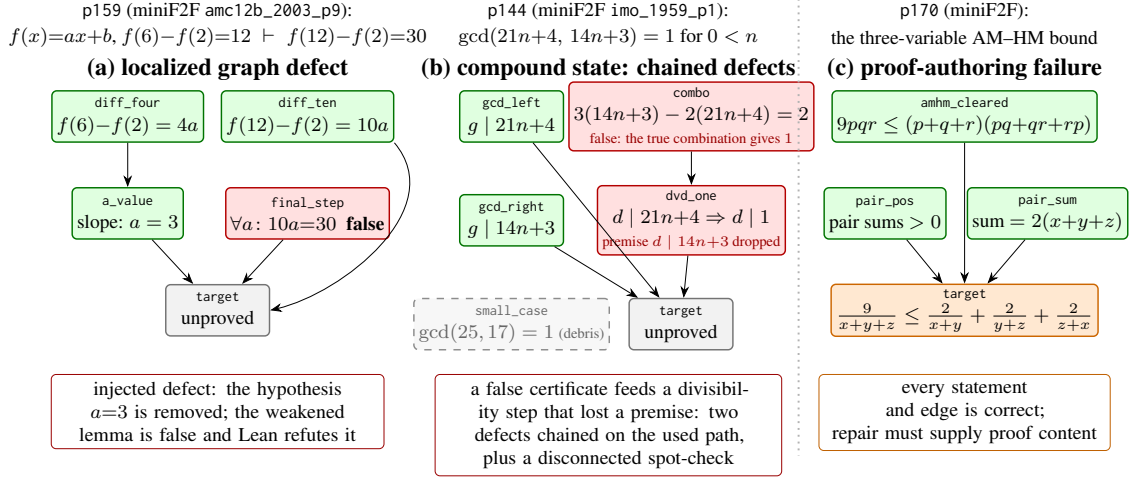

Revising a failed proof plan is not new
\citep{goedelarchitect,editablesketch}; what differs is the editable object: a
declaration-level LeanArchitect graph changed through schema-checked
operations rather than freshly written Lean text. We propose \BR, a typed
local-edit interface for failed Lean blueprints. At
each step, the model chooses one of ten operations that can change a statement,
add or remove an edge, split or delete a node, or write a proof for one node.
The harness applies the operation mechanically and returns Lean feedback. We
compare this interface with two free-form alternatives: atomic
search-and-replace patches over the current source and complete module
rewriting. All three see the same normalized source, typed state, verifier
feedback, model, and budget. The only intended difference is how a repair is
expressed.

A kernel-accepted target theorem is necessary but not sufficient for blueprint
repair. Suppose the proof of a node uses another blueprint lemma but the graph
does not declare that dependency. Lean may still accept the theorem, while the
blueprint remains wrong. We therefore inspect each compiled proof term and
reject any repair with an undeclared blueprint dependency. This check directly
addresses the central distinction of the paper: we evaluate repaired proof
graphs, not only repaired target proofs.

On the $91$ states with localized graph or statement defects, typed repair
solves $79$ states with DeepSeek-V4-Flash \citep{deepseekv4,deepseekv4card} and each
free-form interface solves $81$; a second model, Qwen3.6-Flash
\citep{qwen36card}, shows the same pattern. The small coverage
gap comes with a large efficiency difference. Typed repair reaches $103$ total
solves within $10{,}000$ completion tokens per task, compared with $90$ for
patching and $87$ for rewriting, and has the lowest provider cost per solve.
Patching obtains the highest observed total, though no pairwise difference is
clear of zero, and each interface is the sole solver of a few states, so the
three together cover more than any one of them. These
results support a simple systems view: when a blueprint is mostly correct,
typed local edits recover most of the reachable coverage at the lowest cost,
and free-form generation buys the remainder at a higher price.

\textbf{Our main contributions are:}
\begin{itemize}\itemsep1pt\parskip0pt
\item \textbf{A typed local-repair interface.} Ten schema-checked operations
edit statements, nodes, edges, and proofs while Lean verifies every resulting
state.
\item \textbf{A graph-aware acceptance criterion.} In addition to preserving
and proving the target, an accepted module must declare every inter-node
dependency found in its compiled proof terms.
\item \textbf{A controlled three-interface study.} Typed edits, local source
patches, and whole-module rewrites receive equal source access and feedback,
allowing a direct comparison of coverage, token use, cost, and control.
\item \textbf{A trajectory dataset.} \BT contains the failed states,
construction metadata, accepted and rejected actions, Lean feedback, graph
checks, token and cost records, and final modules.
\end{itemize}

\section{Repair task and benchmark}
\label{sec:problem}

\begin{table}[t]
\centering\small
\caption{Controlled failure families. Excluding the $12$ compound states, the
first five families hold the $91$ edit-shaped states and the last two the $39$
proof-authoring states. A compound state is counted under the family of its
first failing defect but evaluated as a separate stratum. Example repairs are
possibilities, not required routes.}
\label{tab:families}
\begin{tabularx}{\textwidth}{@{}>{\raggedright\arraybackslash}p{0.21\textwidth}rYY@{}}
\toprule
\textbf{Family} & \textbf{$n$} & \textbf{Injected defect} & \textbf{Example typed repair} \\
\midrule
false / too strong & 23 & false statement on the target path & weaken, rewrite, or drop it \\
missing hypothesis & 34 & required assumption removed & strengthen hypotheses \\
missing dependency & 18 & proof uses an undeclared lemma & add the missing edge \\
redundant dependency & 14 & unnecessary node wired into the path & remove edge or node \\
dead node & 14 & node lies outside the target closure & remove the unused node \\
\midrule
monolithic & 31 & intact node not closed by the prover & split or prove the node \\
representation & 8 & inconvenient formal representation & restate and prove \\
\bottomrule
\end{tabularx}
\end{table}

\paragraph{What counts as a repaired blueprint.}
A blueprint is a directed acyclic graph over a Lean module. An edge $u\to v$
means that the proof of node $v$ declares node $u$ as a dependency. Let $C$ be
the target theorem together with every node that can reach it along declared
edges. A state is accepted only when (i) the full module elaborates, (ii) every
node in $C$ is proved, (iii) the target statement is unchanged and contains no
forbidden proof shortcut, (iv) no node is a stored kernel-refuted statement,
and (v) the compiled proofs agree with the declared graph.

The last condition is easiest to state for one proved node $v$.
$\mathrm{Declared}(v)$ is the set of blueprint nodes connected to $v$ by its
incoming \id{proofUses} edges. $\mathrm{Actual}(v)$ is the set of blueprint
nodes reached from the elaborated proof term of $v$; the extractor follows
ordinary helper definitions and stops when it reaches another blueprint node.
We require
\[
  \mathrm{Actual}(v)\subseteq\mathrm{Declared}(v).
\]
Thus a proof may not use an undeclared blueprint lemma. We do not require exact
equality because a repaired proof can stop using an edge that is still
harmlessly declared; we record such unused edges as a separate quality signal.
Nodes outside $C$ must still elaborate and must not be refuted, but they do not
contribute to the target proof and may remain deferred. We separately report
whether every node in the module is complete. Appendix~\ref{app:acceptance}
states the check in full.

\paragraph{Two repair regimes.}
Figure~\ref{fig:regimes} shows the distinction measured by the benchmark. In an
\emph{edit-shaped} failure, at least one statement, node, or edge is wrong. A
proof-only change may bypass the defect, but the failed graph itself contains a
known structural error. In a \emph{proof-authoring} failure, the statements and
edges are intact, but the fixed node prover does not close one of the required
nodes. The label is operational: it describes failure under the fixed prover
used by the harness. Appendix~\ref{app:automation} reports a
stronger-automation baseline on exactly these states.

\paragraph{Controlled states.}
The benchmark contains $142$ failed blueprints over $141$ distinct miniF2F
target theorems \citep{minif2f}. The targets come from the miniF2F Test and
Valid splits at a pinned commit ($244$ items each). A mechanical filter on the
statement's type surface removes $86$ analysis-style items; from the remaining
pool of $402$ we chose targets manually, balancing failure families and topic
areas, and one target contributes two independently constructed states. We
start from correct, author-constructed
blueprints and introduce defects from Table~\ref{tab:families}. Each state is
checked before any model run: false statements have Lean refutations, target
statements are preserved, dependency defects are confirmed against compiled
proof terms, monolithic states keep their original statements and edges, and
representation states swap one statement for a true but inconvenient form. The model is not required to undo the construction edit; any final
module is valid when it proves the original target and passes the acceptance
criterion above.

Of the $142$ states, $91$ contain one localized edit-shaped failure, $39$ are
proof-authoring failures, and $12$ are compound states with linked defects. The
graphs are intentionally small so that the experiment isolates repair rather
than long-horizon search; Table~\ref{tab:shape} gives their size ranges.

\begin{table}[t]
\centering\small
\caption{Graph-size ranges in the controlled benchmark.}
\label{tab:shape}
\begin{tabular}{@{}lrrrr@{}}
\toprule
\textbf{Stratum} & \textbf{$n$} & \textbf{nodes} & \textbf{edges} & \textbf{max depth} \\
\midrule
all & 142 & 1--8 & 0--8 & 3 \\
edit-shaped & 91 & 2--7 & 0--6 & 3 \\
proof-authoring & 39 & 1--5 & 0--4 & 1 \\
compound & 12 & 5--8 & 3--8 & 3 \\
\bottomrule
\end{tabular}
\end{table}

\section{The \BR interface}
\label{sec:interface}

\begin{figure}[b]
\centering
\begin{tikzpicture}[
  font=\scriptsize,
  >={Stealth[length=1.6mm]},
  node distance=2.5mm,
  blk/.style={rounded corners=2pt,draw,semithick,align=center,
    inner sep=2.5pt,text width=0.9\figcol},
]
\node[blk,fill=blue!7,draw=blue!55!black] (state) {failed blueprint};
\node[blk,below=of state,fill=black!4] (llm) {LLM repair policy};
\node[blk,below=of llm,fill=black!4] (tool) {one repair action};
\node[blk,below=of tool,fill=black!4] (harness) {harness applies the action};
\node[blk,below=of harness,fill=blue!7,draw=blue!55!black] (lean)
  {Lean checker and node prover};
\node[blk,below=of lean,fill=blue!7,draw=blue!55!black] (invariant)
  {graph check: every actual proof dependency is declared};
\node[blk,below=of invariant,fill=green!8,draw=green!45!black] (trace)
  {\BT action and feedback log};
\draw[->] (state) -- (llm);
\draw[->] (llm) -- (tool);
\draw[->] (tool) -- (harness);
\draw[->] (harness) -- (lean);
\draw[->] (lean) -- (invariant);
\draw[->] (invariant) -- (trace);
\draw[->] (invariant.west) .. controls +(-0.65,0) and +(-0.65,0) ..
  node[left=1pt,align=center,font=\tiny]{typed\\feedback} (llm.west);
\end{tikzpicture}
\caption{The repair loop used by all three interfaces. The harness applies one candidate action, Lean checks the resulting module, and the graph check returns structured feedback for the next step.}
\label{fig:loop}
\end{figure}
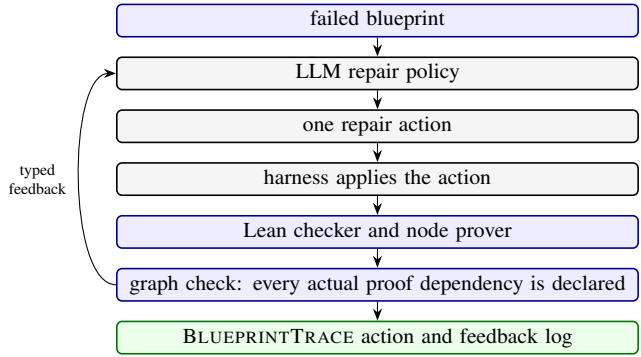

Figure~\ref{fig:loop} shows the complete interaction loop. The model receives
the current blueprint, the current Lean source, and typed diagnostics. It emits
one action in the format required by its interface. The harness applies the
action, elaborates the module, runs the fixed node prover on deferred nodes,
checks actual proof dependencies against declared edges, and returns a short
reason for every accepted or rejected step. The new state then becomes the
input to the next turn. All transitions are stored in \BT.

\paragraph{Typed local edits.}
Our proposed interface exposes the ten operations in
Table~\ref{tab:tools}. A call must satisfy a JSON schema and refer to existing
nodes. The harness rejects unknown names, cycles, forbidden constructs, target
statement changes, and inconsistent proof dependencies. The operations are
local: source outside the requested edit is preserved mechanically.

\begin{table}[t]
\centering\small
\caption{The ten typed operations, grouped by purpose.}
\label{tab:tools}
\begin{tabularx}{0.86\textwidth}{@{}p{0.26\textwidth}Y@{}}
\toprule
\textbf{Purpose} & \textbf{Operations} \\
\midrule
change a statement & \id{weaken\_node}, \id{strengthen\_hypotheses},
\id{rewrite\_node\_statement} \\
change graph structure & \id{split\_node}, \id{add\_dependency},
\id{drop\_dependency} \\
remove a node & \id{drop\_false\_node}, \id{drop\_dead\_node} \\
write proof content & \id{set\_node\_proof} \\
end the episode & \id{stop\_unrepairable} \\
\bottomrule
\end{tabularx}
\end{table}

\paragraph{Two free-form comparisons.}
The local-patch interface returns one or more exact search-and-replace blocks.
All blocks are applied atomically, so a failed match leaves the module
unchanged. The rewrite interface returns a complete replacement module. Patch
and rewrite candidates pass the same parser, allowed-import, namespace,
forbidden-token, target-signature, Lean, and graph-dependency checks as typed
repairs. Table~\ref{tab:interfaces} summarizes what remains different. In
particular, patching controls for the main concern with a rewrite-only
baseline: it sees the full source and can make a small free-form edit without
regenerating untouched code.

\begin{table}[b]
\centering\small
\caption{Interface properties. All three interfaces see the same current
source, typed diagnostics, and verifier feedback.}
\label{tab:interfaces}
\begin{tabularx}{0.86\textwidth}{@{}Yccc@{}}
\toprule
\textbf{Property} & \textbf{Typed} & \textbf{Patch} & \textbf{Rewrite} \\
\midrule
output unit & one operation & edit blocks & full module \\
untouched source preserved & mechanically & mechanically & no \\
semantic action schema & yes & no & no \\
target guard fires & before applying & after applying & after applying \\
\bottomrule
\end{tabularx}
\end{table}

\paragraph{Matched evaluation protocol.}
The main experiment gives every interface the same normalized module, typed
state, verifier feedback, pinned DeepSeek-V4-Flash model, output cap, and
interaction budget. Each step may use one model response; the limit is eight
steps for ordinary states and twelve for compound states. A candidate counts
as solved only after the online acceptance checks described in
Section~\ref{sec:problem}. Appendix~\ref{app:configuration} gives the
configuration and the three system prompts in full; the schemas, Lean versions,
token accounting, and pricing snapshots are part of the artifact.

\paragraph{A second model.}
We repeat the same evaluation with Qwen3.6-Flash. The states,
prompts, interfaces, budgets, output cap, and acceptance checks are unchanged.

\section{Results}
\label{sec:results}

\subsection{Coverage}
Table~\ref{tab:main} reports the controlled benchmark. A state counts as solved
only if the repair is accepted during the run and the stored file passes the
dependency check we repeat afterwards (end of this subsection). With
DeepSeek-V4-Flash, typed local repair solves $79$ of the $91$ edit-shaped
failures, only two fewer than either free-form interface. Patching and
rewriting solve $10$ of the $12$ compound states, and typed repair solves $9$.
On the $39$ proof-authoring failures, local patching leads with
$18$, typed repair solves $16$, and rewriting solves $13$. Across all $142$
states, patching has the highest final coverage ($109$), followed by typed
repair and rewriting ($104$ each).

These gaps are small relative to the sample. On the primary endpoint, the $91$
edit-shaped states, typed repair and patching disagree on four states only: a
paired difference of $-2.2$ points with a $95\%$ interval of $[-6.6,+2.2]$
(exact McNemar $p=0.63$, source theorems resampled), and no pairwise difference
on any stratum is clear of zero. The interval covers uncertainty over states of
this construction, not over repeated samples from the model, and an interval
covering zero is not evidence of equivalence.

\begin{table}[t]
\centering\small
\caption{Solved controlled states. A state counts only if the repair is accepted
during the run and the stored file passes the same later dependency check. Both
models see the same states and are counted the same way.}
\label{tab:main}
\begin{tabular}{@{}lrrrr@{}}
\toprule
\textbf{State type} & \textbf{$n$} & \textbf{Typed} & \textbf{Patch} & \textbf{Rewrite} \\
\midrule
\multicolumn{5}{@{}l}{\emph{DeepSeek-V4-Flash}} \\
edit-shaped & 91 & 79 & 81 & 81 \\
proof-authoring & 39 & 16 & 18 & 13 \\
compound & 12 & 9 & 10 & 10 \\
total & 142 & 104 & 109 & 104 \\
\midrule
\multicolumn{5}{@{}l}{\emph{Qwen3.6-Flash}} \\
edit-shaped & 91 & 70 & 71 & 72 \\
proof-authoring & 39 & 10 & 7 & 7 \\
compound & 12 & 4 & 10 & 5 \\
total & 142 & 84 & 88 & 84 \\
\bottomrule
\end{tabular}
\end{table}

The overlap is also informative. Typed repair and patching solve $78$
edit-shaped states in common; one is solved only by typed repair and three only
by patching. Typed repair and rewriting solve $77$ in common; two are solved
only by typed repair and four only by rewriting. Thus the methods are not only
different encodings of the same successes: each free-form interface adds a
small number of localized repairs, while typed repair also closes cases that
each of them misses.

Over all $142$ states, $89$ are solved by all three interfaces and $24$ by none;
Table~\ref{tab:perfamily} in Appendix~\ref{app:perfamily} splits both by
injected family. Each interface is the sole solver of a few: two states are
solved only by typed repair, three only by patching and three only by
rewriting. The three interfaces together cover $118$ of the $142$ states.

\paragraph{Repeating the dependency check afterwards.}
During a run, a step is refused if the proof it produces uses a blueprint lemma
for which no edge is declared. That check sees the module the harness builds. We
therefore repeat it on the files we store: every stored repair that can be
elaborated again gets one more Lean run, $317$ in all, comparing the lemmas each
proof actually uses against the edges the blueprint declares. No stored repair
hides a dependency: $104$ of $104$ typed, $109$ of $109$ patch, and $104$ of
$104$ rewrite results pass. The reverse is not true: some repairs keep declared edges that their proofs no
longer use; Appendix~\ref{app:acceptance} reports them.

\paragraph{The second model.}
Qwen3.6-Flash solves fewer states in every interface. The comparison between
interfaces still repeats. On the
$91$ edit-shaped failures the three interfaces stay within two states of each
other ($70$ typed, $71$ patch, $72$ rewrite). Two results do not repeat. On the $12$ compound states typed repair
solves $4$ while patching solves $10$, and on the $39$ proof-authoring states
typed repair leads with $10$ against $7$ for both free-form interfaces. The later
dependency check is again passed by every stored repair: $84$ of $84$ typed, $88$
of $88$ patch, and $84$ of $84$ rewrite.

\subsection{Efficiency under equal budgets}
\label{sec:efficiency}

\begin{figure}[t]
\centering
\begin{minipage}[t]{0.48\textwidth}
\centering
{\small\textbf{DeepSeek-V4-Flash}}\\[1.5pt]
\includegraphics[width=\linewidth]{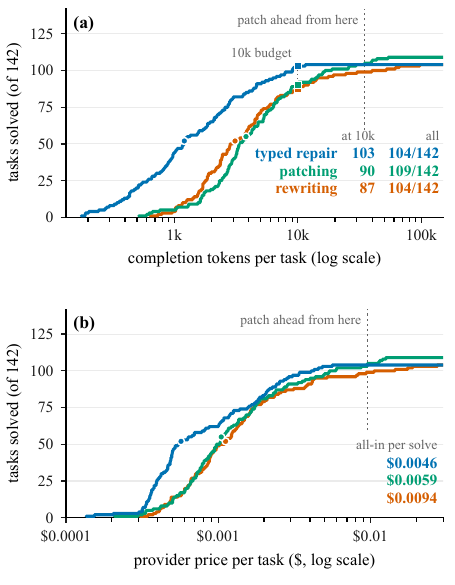}
\end{minipage}\hfill
\begin{minipage}[t]{0.48\textwidth}
\centering
{\small\textbf{Qwen3.6-Flash}}\\[1.5pt]
\includegraphics[width=\linewidth]{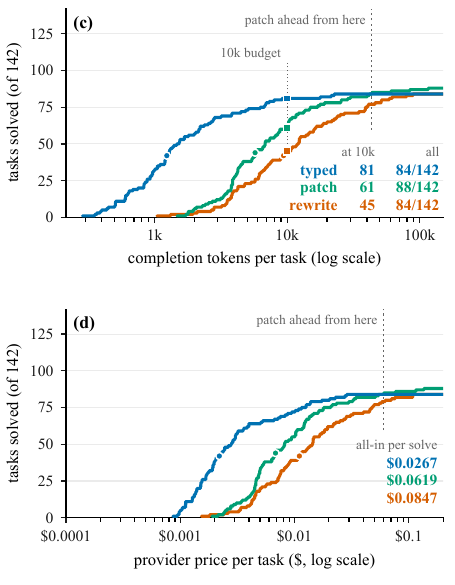}
\end{minipage}
\caption{Cumulative solved states as the per-task budget increases, for both
models. Panels (a) and (c) use completion tokens; panels (b) and (d) use
provider price. All four use the same $142$ initial states, count solved as in
Table~\ref{tab:main}, and stop charging a task when it is first solved. The
token axes are identical, so the two models' left-hand panels can be read
against each other; the price axes are not, because Qwen3.6-Flash output is
priced an order of magnitude above DeepSeek-V4-Flash's.}
\label{fig:frontier}
\end{figure}

Panels (a) and (b) of Figure~\ref{fig:frontier} report the matched
DeepSeek-V4-Flash run. The typed
interface reaches useful coverage earlier: with at most $10{,}000$ completion
tokens per task it has already solved $103$ states, compared with $90$ for
patching and $87$ for rewriting. This is almost the full typed total of $104$.
Patching eventually reaches $109$ and overtakes the typed curve only after
$34{,}767$ completion tokens per task. Rewriting reaches the same total of $104$
as typed repair, but needs $96{,}944$ completion tokens per task to get there.

The same pattern appears in provider cost. Typed repair has the lowest all-in
cost per accepted solve: relative to it, patching is $1.30\times$ as expensive
and full rewriting is $2.06\times$ as expensive. The comparison includes
unsuccessful episodes and all input and output charges. The result is not merely
that typed responses are shorter: the frontier asks how many states are solved
under the same cumulative budget and therefore combines response length, number
of attempts, failures, and early stopping.

\paragraph{The second model.}
Panels (c) and (d) repeat the reading for Qwen3.6-Flash. The ordering between
interfaces is the same and the gaps are wider. Within $10{,}000$ completion
tokens per task, typed repair has already solved $81$ of the $84$ states it
ever solves, against $61$ of $88$ for patching and $45$ of $84$ for rewriting.
Per accepted solve, patching costs $2.32\times$ and rewriting $3.17\times$ as
much as typed repair (\$0.027 per solve for typed repair, \$0.062 and \$0.085
for the free-form interfaces).

\subsection{Control and repair behavior}
All three interfaces keep the target statement, but they stop a change at
different points. A typed operation names the node it edits, so an attempt on
the target is refused before anything is applied; three such calls occur in the
matched run and all are refused. Patching and rewriting write Lean text and are
checked afterwards against the target signature: rewriting returns a module in
which no node carries the target statement in $12$ responses, all rejected, and
every patch keeps the target statement in this run. The graph-dependency gate
also rejects two typed, five patch, and one rewrite candidates that would
otherwise leave proof use inconsistent with the declared blueprint; as reported
above, none of the accepted results violates that rule either.

The interfaces preserve source at different granularities. A typed operation
changes only the requested graph object, and a patch changes only text matched
by its blocks. Rewriting may replace the whole decomposition. We treat graph
size, disconnected nodes, unused edges, statement changes, and full-module
completion as separate quality dimensions rather than collapsing them into the
solve label; Appendix~\ref{app:footprint} reports what each interface leaves
behind.

\paragraph{The remaining gap is not a missing operation.}
Patching solves $11$ states that typed repair does not. Either the typed belt
lacks an operation those repairs need, or it has the operations and the model
chose badly. To tell these apart we take each accepted patch, write by hand the
typed calls that produce the same repair, and run them through the same harness
and the same checks; no model is involved. All $11$ repairs can be expressed: $10$
reproduce the final module exactly, and the eleventh differs only in
declaration names, which no typed operation changes. The longest takes $6$ calls, every program fits the budget
its own episode had, and every intermediate state passes the checks. The gap is
therefore in which actions were chosen, not in which actions exist. This shows
that the repairs are reachable, not that the model would have found them. Both
interfaces saw the same source, and no program reuses a proof body the typed
side was not shown.

\paragraph{Combining the interfaces.}
Replaying the runs we already have in a fixed order (each interface starting
again from the initial failed state, a task stopping at its first accepted
repair) reaches $118$ of the $142$
states. Every order reaches that number, but starting with the typed interface
is the cheapest ($\$1.41$ against $\$1.66$ for the most expensive order).

\section{\BT: states, trajectories, and audits}
\label{sec:dataset}

\BT is the data counterpart of the repair harness. The artifact holds all
$142$ controlled states as a single table: the source theorem, the initial Lean
module, the graph, typed node statuses, and construction metadata. The metadata
records the correct source blueprint or the documented injected defects and the
Lean checks that support the assigned family. This information explains how the
state was created; it does not prescribe the repair that a model must produce.

For each interface, an episode stores every model response, whether it was
applied, the exact graph and source change, Lean and node-prover feedback, the
actual-versus-declared dependency scan, token and cost usage, and the final
module. Rejected actions are retained with their reasons. These rows show which
repair decisions were attempted but invalid in a particular state: for
example, an edge that would create a cycle, a proof that uses an undeclared
node, or a rewrite that changes the target. A release containing only
successful final proofs could not recover these decisions later.

The artifact also contains the exact system prompts, the machine-readable JSON
schemas for all ten typed operations, the pinned Lean environment, the model
identifiers, pricing
snapshots, and scripts that regenerate the reported tables and frontier curves.
This makes the interface contract and the evaluation predicate inspectable,
not just the final solve labels.

\begin{table}[t]
\centering\small
\caption{\BT at a glance.}
\label{tab:trace}
\begin{tabularx}{0.92\textwidth}{@{}p{0.24\textwidth}Y@{}}
\toprule
controlled states & 142 failed blueprints over 141 miniF2F targets; seven
families plus compound states \\
per-step records & source, graph, action, applied change, Lean feedback,
dependency scan, tokens, and cost \\
terminal records & final module, target certificate, strict completion, graph
quality, and solve label \\
reproducibility & exact prompts, all ten schemas, model identifiers, the
pinned Lean environment, and analysis scripts \\
\bottomrule
\end{tabularx}
\end{table}

\section{Related work}
\label{sec:related}

\paragraph{Editable plans and blueprints.}
EditableSketch and SketchRefine preserve proved subgoals while correcting or
further decomposing a proof sketch \citep{editablesketch}. This is the closest
motivation to our local-repair view. The editable object differs: \BR operates
on a LeanArchitect declaration-level DAG and can change formal statements,
proofs, nodes, and explicit dependency metadata. Our additional focus is a
schema-checked action interface, controlled graph-defect families, an
actual-versus-declared proof invariant, and trajectories that retain rejected
edits. Goedel-Architect, LeanMarathon, LEAP, and difficulty-aware decomposition
also revise proof plans with verifier feedback
\citep{goedelarchitect,leanmarathon,leap,quarry}.

\paragraph{Proof repair and verifier feedback.}
Baldur and APOLLO repair proof text from previous attempts and compiler
messages \citep{baldur,apollo}; APRIL and OProver turn failed attempts into
supervision or iterative context \citep{april,oprover}. Proof Repair across
Type Equivalences transforms Coq proof terms after type changes
\citep{proofrepair}; our task instead repairs a failed Lean blueprint and its
dependency metadata. VERITAS and process-verified reinforcement learning use
Lean feedback for search or credit assignment \citep{veritas,pvrl}.

\paragraph{Process data, cost, and benchmark checks.}
FormalRewardBench injects controlled proof errors to evaluate reward models
\citep{formalrewardbench}; \BT adds stateful graph actions and typed rejection
reasons. Cost-aware agent control chooses whether to continue or restart a
proof plan \citep{costquality}; our frontier instead compares three repair
interfaces under matched budgets. Recent benchmark audits show why a
kernel-correct target alone is not a complete evaluation guarantee
\citep{formalbenchmarkfaults}. This motivates our preserved target provenance,
explicit graph-dependency check, and retained audit records.

\section{Conclusion}
\label{sec:conclusion}

Agents already repair code through a loop of tool calls. \BR asks what changes
when the tools are ten typed operations on a proof graph rather than free-form
edits to a file, and when a repair is accepted only if its compiled proof agrees
with the declared graph. With DeepSeek-V4-Flash, on $91$ localized failures,
typed edits solve $79$ states versus $81$ for both local patching and complete
rewriting. Across all $142$ states, patching reaches the highest observed
total (no pairwise coverage difference is clear of zero), but
typed repair reaches almost all of its coverage much earlier: $103$ solves at
a $10{,}000$-completion-token budget, compared with $90$ and $87$. It also has
the lowest cost per solve; patching is $1.30\times$ as expensive and rewriting
$2.06\times$ as expensive. A second model, Qwen3.6-Flash, solves fewer states and keeps
the same cost ordering, although its coverage pattern breaks on the states that
carry more than one defect. Finally, typed repair refuses a target
statement change before it is applied, and every accepted endpoint is checked for
undeclared proof dependencies. \BT exposes the full interaction data needed to
learn better repair policies from both successful and rejected structural
edits.

\section{Limitations}

\paragraph{Small controlled graphs.}
The benchmark isolates local repair on blueprints with one to eight nodes and
target paths of depth at most three. The compound set contains only twelve
states. Results may change on research-scale formalizations with longer paths,
more shared definitions, and defects introduced over many refinement rounds.

\paragraph{Stochastic model behavior.}
The reported table uses one episode per state and interface, so it describes
the observed runs rather than an expectation over repeated samples. Repeating
the same three-way protocol with a second model tests whether the qualitative
pattern transfers across model families, but it does not replace multi-seed
estimation within one model.

\paragraph{Operational proof-authoring label.}
A state is called proof-authoring when the fixed
\id{rfl}/\hspace{0pt}\id{simp}/\hspace{0pt}\id{omega}
node prover does not close an otherwise intact graph: no statement is false
and no dependency is wrong. The label describes the
harness and budget, not intrinsic mathematical difficulty. Stronger automation
does not remove the regime: five standard tactics close $3$ of the $39$ states,
and only one of those is a state that no interface solved.

\paragraph{Implemented interfaces are bundles.}
Typed repair, local patching, and rewriting differ in several linked ways:
action schema, output granularity, preservation of untouched source, and how
target immutability is enforced. A patch reply may contain several
search-and-replace blocks; a typed turn applies exactly one operation. Equal
step budgets therefore do not mean equal numbers of small edits; the token and
cost comparisons include this difference and do not isolate it. The experiment
compares these complete
interfaces. It does not isolate a causal effect of typing alone, although the
local-patch arm separates locality from complete regeneration more directly
than a rewrite-only comparison.

\paragraph{Tool vocabulary and training.}
The ten operations are one practical vocabulary, not a proof that no better
granularity exists. The evaluated models were not trained to use it. Several
mathematical repairs can be expressed by multiple action sequences, so future
training should allow multiple valid routes and use rejected-action reasons as
additional supervision.

\paragraph{Benchmark provenance.}
Targets come from miniF2F and may have appeared in model training data. Correct
blueprints and injected defects are author-constructed and reviewed with
mechanical Lean checks, but there is no second independent annotator. \BT is a
controlled diagnostic benchmark, not a sample of failures from a real
generation pipeline. The results say where typed edits work well (on
mostly-correct blueprints with localized defects), not that typed edits win on
arbitrary blueprint failures.

\clearpage
\appendix
\begin{center}
{\Large\bfseries Appendices}
\end{center}
\vspace{0.5em}

\section{A representative split-node repair}
\label{app:examples}

\subsection{Splitting a hard induction node}
\label{app:p40}

State \id{p40} asks to prove that, for every natural number $n$,
\[
  12 \mid 4^{n+1}+20.
\]
The failed blueprint contains this fact as one monolithic, unproved target node.
The fixed node prover does not close it, although the statement is correct. The model therefore calls \id{split\_node} and turns the single node
into a small induction blueprint.

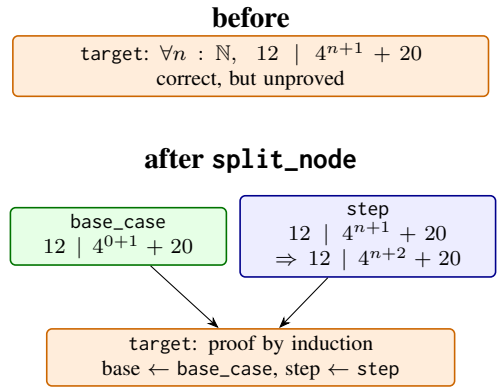
\begin{figure}[htbp]
\centering
\begin{tikzpicture}[
  font=\scriptsize,
  >={Stealth[length=1.6mm]},
  n/.style={rounded corners=2pt,draw,semithick,align=center,inner sep=2.4pt},
  hard/.style={n,draw=orange!80!black,fill=orange!14,text width=0.82\figcol},
  ok/.style={n,draw=green!45!black,fill=green!10,text width=0.36\figcol},
  helper/.style={n,draw=blue!55!black,fill=blue!7,text width=0.43\figcol},
]
\node[font=\bfseries] at (0,0.65) {before};
\node[hard] (before) at (0,0)
  {\id{target}: $\forall n\!:\!\mathbb{N},\;12\mid 4^{n+1}+20$\\
   correct, but unproved};

\node[font=\bfseries] at (0,-1.25) {after \id{split\_node}};
\node[ok] (base) at (-1.75,-2.25)
  {\id{base\_case}\\$12\mid 4^{0+1}+20$};
\node[helper] (step) at (1.55,-2.25)
  {\id{step}\\$12\mid 4^{n+1}+20$\\$\Rightarrow 12\mid 4^{n+2}+20$};
\node[hard,text width=0.70\figcol] (target) at (0,-3.85)
  {\id{target}: proof by induction\\base $\leftarrow$ \id{base\_case},
   step $\leftarrow$ \id{step}};
\draw[->] (base) -- (target);
\draw[->] (step) -- (target);
\end{tikzpicture}
\caption{The \id{p40} repair. A monolithic induction goal is replaced by a
base case and an induction step.}
\label{fig:p40}
\end{figure}

The split is structurally correct, but the induction step still needs a proof.
The complete five-call trajectory is:

\begin{center}\small
\begin{tabularx}{\textwidth}{@{}c >{\raggedright\arraybackslash}p{0.28\textwidth}Y@{}}
\toprule
\textbf{Step} & \textbf{Operation} & \textbf{Lean feedback / result} \\
\midrule
1 & \id{split\_node} & refused before anything is applied: the proposed proof
uses \id{native\_decide} \\
2 & \id{split\_node} & applied; \id{base\_case} closes at once, and the proof
offered for \id{step} does not elaborate \\
3 & \id{rewrite\_node\_\hspace{0pt}statement} on \id{step} & applied; a second proof for the
same statement, still not elaborating \\
4 & \id{set\_node\_proof} on \id{step} & \id{step} proved by \id{omega} \\
5 & \id{set\_node\_proof} on the target & the target closes by induction; the
state is solved \\
\bottomrule
\end{tabularx}
\end{center}

This example shows that typed repair is not limited to deleting or rewiring
nodes: \id{split\_node} can introduce a useful proof decomposition, after which
local proof edits complete it. The refused first call is ordinary rather than
exceptional: the screen that protects the target statement also rejects
forbidden proof constructs before a candidate reaches Lean.

\section{Acceptance and benchmark details}
\label{app:acceptance}

\subsection{Graph-aware terminal acceptance}
For the target closure $C$ defined in Section~\ref{sec:problem}, the harness
accepts an assembled module only if all of the following checks pass:
\begin{enumerate}\itemsep1pt\parskip0pt
\item The complete module elaborates in a fresh Lean process without an error
or timeout.
\item No blueprint node matches a stored kernel refutation certificate.
\item Every node in $C$ is proved; proofs produced by the fixed node prover are
spliced into the module and the assembled module is elaborated again.
\item The target statement hash equals the original and its axiom certificate
contains no \id{sorryAx}, \id{native\_decide}, or nonstandard axiom.
\item For every node with a real proof,
$\mathrm{Actual}(v)\subseteq\mathrm{Declared}(v)$. The extractor follows
non-blueprint constants and stops at blueprint declarations.
\end{enumerate}
Conditions 1, 2, and 5 range over the whole module. Only proof completion is
restricted to $C$, so a disconnected deferred node does not make the target
proof invalid; strict all-node completion is reported separately.

Condition 5 is an inclusion, not an equality: a repaired proof may stop using an
edge that stays harmlessly declared. With DeepSeek-V4-Flash such edges remain on
$10$ typed, $28$ patch and $6$ rewrite results. On the typed side these are $12$
edges, and $8$ of them belong to nodes closed by the fixed node prover rather
than to proofs the model wrote. We report unused edges as a quality measure and
require only that no dependency stays hidden.

\subsection{Stronger automation}
\label{app:automation}
The fixed node prover tries \id{rfl}, then \id{simp}, then \id{omega}, with
$200{,}000$ heartbeats for each tier; six states also declare \id{norm\_num} as
a fourth tier. On the $39$ proof-authoring states we ran each of \id{norm\_num},
\id{ring}, \id{linarith}, \id{nlinarith}, and \id{aesop} on its own at the same
budget, counting only proofs the kernel accepts without extra axioms. Two
states close with \id{aesop} and one with \id{nlinarith}; the other three
tactics close none. Six modules needed one added import for these tactics to be
in scope at all.

\subsection{Outcomes by failure family}
\label{app:perfamily}
Table~\ref{tab:perfamily} splits the controlled benchmark by injected family.
A compound state carries more than one defect, so it is counted once, in its
own row, and not inside the family of its first defect. Redundant dependencies,
missing dependencies and false lemmas are solved by all three interfaces almost
without exception. The states that no interface solves sit in two families:
$15$ of the $24$ are monolithic nodes and $6$ are missing hypotheses.

\begin{table}[htbp]
\centering\small
\caption{Solved states by failure family, for DeepSeek-V4-Flash, on the same
endpoints as Table~\ref{tab:main}.}
\label{tab:perfamily}
\begin{tabular}{@{}lrrrrr@{}}
\toprule
\textbf{Family} & \textbf{$n$} & \textbf{Typed} & \textbf{Patch} &
\textbf{Rewrite} & \textbf{None} \\
\midrule
false / too strong   &  18 &  17 &  16 &  18 &  0 \\
missing hypothesis   &  30 &  21 &  24 &  21 &  6 \\
missing dependency   &  15 &  15 &  15 &  15 &  0 \\
redundant dependency &  14 &  14 &  14 &  14 &  0 \\
dead node            &  14 &  12 &  12 &  13 &  1 \\
monolithic           &  31 &  12 &  12 &  11 & 15 \\
representation       &   8 &   4 &   6 &   2 &  2 \\
compound             &  12 &   9 &  10 &  10 &  0 \\
\midrule
total                & 142 & 104 & 109 & 104 & 24 \\
\bottomrule
\end{tabular}
\end{table}

\subsection{Footprint of the accepted repairs}
\label{app:footprint}
The interfaces differ in what they leave behind, on the same
DeepSeek-V4-Flash endpoints as Table~\ref{tab:main}. Every accepted result of
every interface is complete in the strict sense: no node is left unfinished
anywhere in the module, not only inside the target closure. Rewriting most
often returns a blueprint holding the target statement alone, on $42$ of $104$
results; typed repair does so on $38$ of $104$ and patching on $14$ of $109$.
Patching leaves nodes outside the target closure on $26$ results, typed repair
on $19$, and rewriting on $5$. Patching also changes node statements most
often: $62$ changed statements against $16$ for typed repair and $21$ for
rewriting.

\section{Experimental configuration}
\label{app:configuration}
All model calls go through OpenRouter. The matched run uses DeepSeek-V4-Flash
build
\id{deepseek-v4-flash-\hspace{0pt}20260423} \citep{deepseekv4card}; the repeated run uses
Qwen3.6-Flash \citep{qwen36card}, model identifier \id{qwen/qwen3.6-flash},
served by a single endpoint (Alibaba).
Neither run sends decoding
parameters, so provider defaults apply. The maximum model output is $49{,}152$
tokens in every arm. The three interfaces together cost \$2.10 with
DeepSeek-V4-Flash and \$14.80 with Qwen3.6-Flash. The artifact pins the
Lean/mathlib environment and records input, cached input, reasoning, and output
tokens together with the provider pricing snapshot.

\subsection{The prompts as sent}
The three system prompts follow, verbatim, for state \id{p02}; across states
only the namespace, the label prefix and the allowed imports differ. The typed
arm also receives the ten operation schemas in the request's function-calling
field, contained in full in the artifact.
\paragraph{Typed local edits.}
\begin{Verbatim}[fontsize=\footnotesize,breaklines,breakautoindent=false,breaksymbolleft={},xleftmargin=0pt]
You repair formal proof blueprints (Lean 4, LeanArchitect @[blueprint] attribute). A blueprint is a DAG of theorem nodes; the state below FAILED for the typed reasons given. You control the blueprint ONLY through the provided tools: each turn call EXACTLY ONE tool; the harness applies it mechanically to the Lean module, runs the Lean verifier and a deterministic node prover (rfl, simp, omega on deferred nodes), and returns typed feedback. Iterate until the target is proved or the budget runs out.

Rules:
- The target theorem's statement is IMMUTABLE (its proof may change).
- Proofs may be deferred with 'sorry' / 'sorry_using [deps]': the prover tries each deferred node and proved nodes are assembled automatically — a good architecture of simple deferred steps wins without written proofs.
- 'axiom', 'admit', 'native_decide' are forbidden and rejected.
- A rejected tool call leaves the state unchanged but consumes budget; the error message explains why.
- The graph in each tool result is the CURRENT state after your edit.
Plan your edits from the typed statuses: statement_refuted means the node is kernel-refuted (drop or replace it); declaration_uses_sorry means the proof is deferred; no_proof_within_budget / budget_ladder_exhausted mean the prover failed on it as stated (split it or prove it yourself).
\end{Verbatim}
\paragraph{Local patching.}
\begin{Verbatim}[fontsize=\footnotesize,breaklines,breakautoindent=false,breaksymbolleft={},xleftmargin=0pt]
You repair formal proof blueprints (Lean 4 with the LeanArchitect @[blueprint] attribute). A blueprint is a DAG of theorem nodes; the state below FAILED for the typed reasons given. You edit the CURRENT Lean module source directly with exact search/replace patches. Each turn reply with ONE patch attempt: one or more blocks in EXACTLY this format (only the blocks are interpreted):

<<<<<<< SEARCH
(text that occurs in the current module source)
=======
(replacement text)
>>>>>>> REPLACE

Patch semantics (mechanical, checked before evaluation):
- SEARCH must be non-empty and must match the current module source EXACTLY ONCE, character for character, whitespace and line breaks included. To insert, anchor on neighboring existing text and repeat it in the replacement; to delete, leave the replacement side empty.
- Blocks apply in reply order, each against the text produced by the previous block. The patch is ATOMIC: the first failing block rejects the whole attempt and the module stays unchanged. A rejected attempt still consumes budget; the error message explains why.
- After a patch applies, the harness validates the module, elaborates it with the Lean verifier, runs the deterministic node prover on deferred nodes, and returns typed feedback plus the UPDATED module source. Write every SEARCH against the LATEST module source shown to you.

Rules:
- The target theorem's statement (binders and result type) must be preserved EXACTLY; a patch whose result changes it is rejected after application.
- 'axiom', 'admit', 'native_decide' are forbidden and rejected.
- The module must keep `namespace P02Bad` ... `end P02Bad`, its imports within: Architect, Mathlib.Tactic.Ring, Mathlib.Tactic.Linarith, Mathlib.Tactic.NormNum, and every declaration a theorem carrying an @[blueprint "p02bad-..."] annotation; node-to-node dependencies are declared with (proofUses := [name1, name2]) on the USING node.
- Proofs may be deferred: use `:= by\n  sorry` for a lone node or `:= by\n  sorry_using [dep1, dep2]` when the planned proof will use those blueprint nodes. A deterministic prover (rfl, then simp, then omega; 200000 heartbeats each) will attempt every deferred node, and proved nodes are assembled automatically — a good architecture of simple deferred steps can win without any written proof.

Worked example (form only; the content is yours):

<<<<<<< SEARCH
theorem step_one (n : Nat) : n = n := by
  sorry
=======
theorem step_one (n : Nat) : n = n := by
  rfl
>>>>>>> REPLACE

Iterate until the target is proved or the budget runs out.
\end{Verbatim}
\paragraph{Whole-module rewriting.}
\begin{Verbatim}[fontsize=\footnotesize,breaklines,breakautoindent=false,breaksymbolleft={},xleftmargin=0pt]
You repair formal proof blueprints (Lean 4 with the LeanArchitect @[blueprint] attribute). A blueprint is a DAG of theorem nodes; the state below FAILED for the typed reasons given. Propose a COMPLETE REPLACEMENT blueprint for the same target: any architecture you want (any number of support lemmas and dependencies), as long as the target theorem statement is preserved exactly.

You will iterate: after each attempt the harness elaborates your module with
the Lean verifier, runs the deterministic node prover (rfl, simp, omega) on
deferred nodes, and returns feedback. Reply to feedback with the COMPLETE
corrected module (one lean fence, full contract), not a description.

Output contract (mechanical, checked before evaluation):
- Reply with EXACTLY ONE fenced code block labeled lean containing a complete Lean 4 module; no other code blocks.
- The module imports must be a subset of: Architect, Mathlib.Tactic.Ring, Mathlib.Tactic.Linarith, Mathlib.Tactic.NormNum.
- The module must declare `namespace P02Bad` and `end P02Bad`.
- Every blueprint node is a theorem carrying the attribute
  @[blueprint "<label>" (latexEnv := "lemma")] for support lemmas or
  @[blueprint "<label>"] for the target theorem, where every <label> starts with "p02bad-" (for example "p02bad-target").
- Declare node-to-node dependencies in the attribute with (proofUses := [name1, name2]) on the USING node; dependencies not declared there are invisible to the graph.
- The target theorem's statement text (binders and result type) must be preserved EXACTLY as given; changing it invalidates the candidate.
- Proofs may be deferred: use `:= by\n  sorry` for a lone node or `:= by\n  sorry_using [dep1, dep2]` when the planned proof will use those blueprint nodes. A deterministic prover (rfl, then simp, then omega; 200000 heartbeats each) will attempt every deferred node, and proved nodes are assembled automatically — a good architecture of simple deferred steps can win without any written proof.
- You may also write complete proofs using core Lean lemmas and the allowed imports.

Follow this reply template EXACTLY (structure and syntax; content is yours;
the fence label is `lean`, and `import Architect` must be the first import
— the blueprint attribute lives there):

action: <action name>

```lean
import Architect

namespace P02Bad

@[blueprint "p02bad-step-one"
  (latexEnv := "lemma")]
theorem step_one (n : Nat) : n = n := by
  sorry

@[blueprint "p02bad-target"
  (proofUses := [step_one])]
theorem target_name (n : Nat) : n = n := by
  sorry_using [step_one]

end P02Bad
```
\end{Verbatim}

\section{Recorded trace fields}
\label{app:tracefields}
Each \BT episode records the protocol configuration; initial state and target
signature; canonical source, node statements, statuses, and edges at every
step; the typed call, patch blocks, or rewritten module; whether the action was
applied and its source/graph difference; Lean, node-prover, and graph-scan
feedback; every rejection reason; token and cost usage; and the final outcome.
Controlled states also include source-target provenance, construction metadata,
and an available correct source blueprint or defect manifest; the ten function
schemas ship as one machine-readable file.

\end{document}